\pdfoutput=1
\documentclass[11pt]{article}

\usepackage[]{acl}

\usepackage{times}
\usepackage{latexsym}
\usepackage[final]{graphicx}
\usepackage{subcaption}

\usepackage[T1]{fontenc}
\usepackage[utf8]{inputenc}

\usepackage{multirow}
\usepackage{tablefootnote}

\usepackage{microtype}

\title{Zero-Shot Ranking Socio-Political Texts with Transformer Language Models to Reduce Close Reading Time}

\author{Kiymet Akdemir \\
  Boğaziçi University\\
  \texttt{\small kiymet.akdemir@boun.edu.tr} \\\And
  Ali Hürriyetoğlu \\
  KNAW Humanities Cluster DHLab \\
  \texttt{\small ali.hurriyetoglu@dh.huc.knaw.nl} \\}

\begin{document}
\maketitle

\begin{abstract}
We approach the classification problem as an entailment problem and apply zero-shot ranking to socio-political texts. Documents that are ranked at the top can be considered positively classified documents and this reduces the close reading time for the information extraction process. We use Transformer Language Models to get the entailment probabilities and investigate different types of queries. We find that DeBERTa achieves higher mean average precision scores than RoBERTa and when declarative form of the class label is used as a query, it outperforms dictionary definition of the class label. We show that one can reduce the close reading time by taking some percentage of the ranked documents that the percentage depends on how much recall they want to achieve. However, our findings also show that percentage of the documents that should be read increases as the topic gets broader.
\end{abstract}

\section{Introduction}

For the information retrieval process positively labeled documents in a dataset are important and should not be missed, therefore achieving high recall is extremely important. However, there is generally a large number of documents that are relevant or not to the concerned topic and doing close reading for all documents and annotating them requires lots of time and resources~\cite{Hurriyetoglu+16,Hurriyetoglu+17}. Therefore, ranking documents according to relevance to the investigated class may help to reduce close reading time and decrease the likelihood of missing critical information.

\citet{books/aw/Baeza-YatesR99} propose ranking documents in decreasing order of being relevant to a given query to accelerate the information retrieval process. \citet{halterman-etal-2021-corpus} apply this method with Natural Language Understanding (NLU) models for binary classification problems using the entailment probabilities of a document and a declarative form of the label. Therefore, to catch a high percentage of positively labeled documents, reading some percentage of documents would be enough since documents that are relevant would be at the top with a high probability. However, their dataset India Police Events focuses on a relatively specific task in information retrieval that is police actions like killing, arresting, failing to intervene, etc. Besides, they apply this method at the sentence level and as they also stated their model suffers from understanding multi-sentence context that increases the false negative rate.

We apply this approach to ProtestNews dataset \citep{hurriyetoglu-etal-2021-multilingual} along with the India Police Events dataset \citep{halterman-etal-2021-corpus} and investigate whether sentence level evaluation or document level evaluation ranks positive documents at the higher level measured by different evaluation metrics. We further investigate whether using the dictionary definition of a class or the declarative form of a class for the query performs this task better.  We compare two NLU models DeBERTa-Large-MNLI \citep{https://doi.org/10.48550/arxiv.2006.03654} and RoBERTa-Large-MNLI \citep{https://doi.org/10.48550/arxiv.1907.11692} in terms of recall and mean average precision.

We present the related work in Section \ref{related-work}. Next, we introduce two datasets we used in our experiments in Section \ref{data}. Then we explain our methodology and list all queries used in this work in Section \ref{method}. We detail our experiments for both datasets and present results in Section \ref{experiments}. Finally, we conclude this work in Section \ref{conclusion} and state what can be done as future work in Section \ref{future-work}.

\section{Related Work} \label{related-work}

\paragraph{Protest Event Detection}
Protest event extraction holds an important place in political social sciences and detection of protest events is generally the first step of the extraction. Due to the cost of manual event extraction, besides the presence of digital news articles and enhancing machine learning methods; automated event extraction comes into play.\\
\citet{hanna_2017} presents MPEDS, an automated system for protest event extraction that contains an ensemble of shallow machine learning classifiers (SVM, SGD and Logistic Regression) to detect protest-related documents. \citet{caselli-etal-2021-protest} proposes Domain Adaptive Retraining for Transformer Language Models and shows that further training BERT with domain-specific dataset improves the performance. They present PROTEST-ER by retraining pre-trained BERT with protest related data from TREC Washington Post Corpus. \citet{wiedemann-etal-2022-generalized} classifies protest related documents in German local news using Pretrained Language Models. They attempt to improve performance and generalizability by eliminating protest-unrelated sentences with keyword search and also by masking named entities with the idea of models may overfit on data by recognizing actors, organizatons and places.\\
\citet{elsafoury2019detecting} focuses on both protest events and police actions i.e. protest repression events in Twitter with Machine Learning models with the claim of news articles suffer from bias, censorship and duplication. \citet{won2017protest} detects and analyze protest events in geotagged tweets and associated images with Convolutional Neural Networks.

\paragraph{Ranking Documents with Transformer Language Models}
\citet{10.1145/3404835.3462812} presents a comprehensive survey of how BERT \citep{devlin-etal-2019-bert} works, ranking documents with BERT, retrieve and rerank approach with monoBERT, ranking metrics, etc.  One of the most remarkable works in the survey is monoBERT and duoBERT, a multistage ranking approach with transformer language models proposed by \citet{Nogueira2019MultiStageDR}. The first stage retrieves the candidate documents with BM25 by treating the query as a bag of words and later, documents are reranked with their relevance score with BERT. DuoBERT also takes into account one document being more relevant than the other at a third stage. However, we rank the documents with a language model at one stage.

\citet{halterman-etal-2021-corpus} rank documents with RoBERTa-Large-MNLI \citep{https://doi.org/10.48550/arxiv.1907.11692} on sentence level by being relevant to a police activity. Yet sentence level evaluation does not take into consideration the relationship between the sentences. Moreover, the task of extracting police events is a relatively specific topic in political event extraction. We apply this method with different document sizes and test on datasets in different topic specificities.

\paragraph{Transformer Language Models DeBERTa and RoBERTa}
DeBERTa-Large-MNLI (DLM) \citep{https://doi.org/10.48550/arxiv.2006.03654} and RoBERTa-Large-MNLI (RLM) \citep{https://doi.org/10.48550/arxiv.1907.11692} are pre-trained language models that improve BERT. Both models are pre-trained on Wikipedia (English Wikipedia dump3; 12GB), BookCorpus (6GB), OPENWEBTEXT (38GB), and STORIES (a subset of CommonCrawl (31GB) and fine-tuned for MNLI task. RLM has a token limitation of 512 whereas DLM has a limitation of theoretically 24,528. We limit the inputs to 512 tokens for both models to be able to compare them fairly. \citet{9673585} find that DLM achieves a better performance in different sentence similarity tasks with respect to RLM and BERT. \citet{https://doi.org/10.48550/arxiv.2006.03654} also show that DeBERTa outperforms RoBERTa in a variety of NLP tasks even when DeBERTa is trained on half of the training data. Therefore, we use DLM and compare it with RLM for our task.

\paragraph{Transferring Question Answering to Entailment Problem}
\citet{Khot_Sabharwal_Clark_2018} and \citet{https://doi.org/10.48550/arxiv.1809.02922} transfer the question answering problem to the entailment problem by forming the question into a declarative form. \citet{clark-etal-2019-boolq} transfer yes/no question answering to entailment problem by training supervised models on entailment datasets and treating entailment probabilities as the probability of the answer being yes. They also use pre-trained ELMo, BERT, and OpenAI GPT as unsupervised models and show that fine-tuning BERT on entailment dataset MultiNLI boosts the performance. The problem of any binary classification can be also transferred to an entailment problem similar to the yes/no question answering, by considering the probability of entailment as the probability of data belonging to the positive class.

\section{Data} \label{data}
We carried out the experiments on two different datasets: India Police Events dataset\footnote{Data and code are provided at \url{https://github.com/slanglab/IndiaPoliceEvents}.} \citep{halterman-etal-2021-corpus} and the ProtestNews dataset of the workshop CASE @ ACL-IJCNLP 2021\footnote{Information and data are provided at \url{https://github.com/emerging-welfare/case-2021-shared-task}.
} \citep{hurriyetoglu-etal-2021-multilingual}.

India Police Events dataset includes 1,257 articles about the Indian state Gujarat from The Times of India and from March 2002. The articles are in English and contain 21,391 sentences in total. Each sentence is classified into 5 different labels regarding police activity: kill, arrest, fail, force, and any action. Question form of the each event type is given in Table \ref{tab:indiaQuestion}. A document belongs to a class if any of its sentences belongs to that class. Table \ref{tab:indiaData} illustrates the number of positive documents and the proportion of the positive documents for each event class. Note that one document may belong to one class, several classes or none of them.

\begin{table}
\centering
\begin{tabular}{ll}
\hline
\textbf{Event type} & \textbf{Question}\\
\hline
kill & Did police kill someone? \\
arrest & Did police arrest someone? \\
fail & Did police fail to intervene? \\ 
force & Did police use force or violence? \\ 
any action & Did police do anything? \\\hline
\end{tabular}
\caption{Question form of each event type.}
\label{tab:indiaQuestion}
\end{table}

\begin{table}
\centering
\begin{tabular}{ll}
\hline
\textbf{Event type} & \textbf{Positive Documents}\\
\hline
kill & 50 (3.98\%) \\
arrest & 128 (10.17\%) \\
fail & 114 (9.05\%) \\ 
force & 90 (7.15\%) \\ 
any action & 457 (36.24\%) \\\hline
\end{tabular}
\caption{Number of positive documents for each event class (India Police Events Dataset).}
\label{tab:indiaData}
\end{table}

\begin{table}
\centering
\begin{tabular}{lll}
\hline
\textbf{Dataset} & \textbf{Positive Documents}\\
\hline
ProtestNews Dataset & 1,912 (20.51\%) \\
ProtestNews Subset & 268 (21.32 \%) \\\hline
\end{tabular}
\caption{Number of positive documents for ProtestNews Dataset and its subset.}
\label{tab:protestData}
\end{table}

ProtestNews dataset includes local news articles of countries India, China, Argentina, and Brazil. These articles are in English, Spanish, Portuguese, and Hindi. For this work, we have only used English articles. There are 9,327 English documents but to equalize data sizes with the India Police Events Dataset we randomly selected 1,257 articles among those. Documents that contain past or ongoing protest events are labeled as positive \citep{https://doi.org/10.48550/arxiv.2206.10299}. Number and proportion of positive documents are given in Table \ref{tab:protestData}.

\section{Method} \label{method}

First, the probability of entailment for each document and a query is calculated with NLU models from Huggingface\footnote{\url{http://huggingface.co}}, and documents are ranked by the decreasing probability of being relevant to the query. Thus we expect the documents that are more relevant are ranked at the top.

Entailment probabilities are evaluated on both sentence and document levels. At sentence level evaluation, entailment probabilities of sentences in a document with the given query are calculated and the largest probability among the sentences is considered as the probability of the document being relevant. For the document level evaluation since RLM is limited to 512 tokens, we divided documents into parts such that each part does not exceed 512 tokens. Similar to the sentence-level approach, probabilities of each part are calculated and the one with the largest probability is considered as the probability of the document. After getting the probabilities for all documents, they are ranked in the decreasing probability.

We compare the results by checking how much recall is achieved when a specified proportion of data is read from the ranked documents following Halterman et. al. (2021) and also by calculating the mean average precision. We release our code publicly\footnote{\url{https://github.com/kiymetakdemir/zero-shot-entailment-ranking}}.

\subsection{Models}
We focused on the performances of two multilingual NLU models that are RLM\footnote{\url{https://huggingface.co/roberta-large-mnli}} \citep{https://doi.org/10.48550/arxiv.1907.11692} and DLM\footnote{\url{https://huggingface.co/microsoft/deberta-large-mnli}} \citep{https://doi.org/10.48550/arxiv.2006.03654} which are pre-trained on the same datasets (Wikipedia and BookCorpus). We conduct experiments with both models and compare the results.

\begin{table*}[h!]
    \centering
    \begin{tabular}{p{6cm} p{8cm}}
    \hline
    \textbf{Query type} & \textbf{Query}\\
    \hline
      Declarative  query  & There is a protest.\\
      \hline
      Definitional query & There is a strong complaint expressing disagreement, disapproval, or opposition. (definition of protest\tablefootnote[9]{\url{https://dictionary.cambridge.org/dictionary/english/protest}})\\
      \hline
      Social protest definition (Annotation Manual) & Individuals, groups, or  organizations voice their objections, oppositions, demands or grievances to a person or institution of authority. \\
      \hline
      Contentious politics event definition (Annotation Manual) & There is a politically motivated collective action event.\\
     \hline
     'protest' + definitional query & Protest, there is a strong complaint expressing disagreement, disapproval, or opposition.\\
     \hline
     Protest definition + opposition definition & There is a strong complaint expressing disagreement, disapproval, or opposition. Disagreement with something, often by speaking or fighting against it, or (esp. in politics) the people or group who are not in power. (definition of opposition\tablefootnote[10]{\url{https://dictionary.cambridge.org/dictionary/english/opposition}})\\
     \hline
     Protest definition + disapproval definition & There is a strong complaint expressing disagreement, disapproval, or opposition. The feeling of having a negative opinion of someone or something. (definition of disapproval\tablefootnote[11]{\url{https://dictionary.cambridge.org/dictionary/english/disapproval}})\\
     \hline
    \end{tabular}
    \caption{Queries used for the ProtestNews dataset.}
    \label{tab:protestQuery}
\end{table*}

\begin{table*}[t]
\centering
\begin{tabular}{llp{8cm}}
\hline
\textbf{Event type} & \textbf{Declarative query} & \textbf{Definitional query}\\
\hline
kill & Police killed someone. & Police caused someone or something to die. (definition of kill\tablefootnote[12]{\url{https://dictionary.cambridge.org/dictionary/english/kill}}) \\
\hline
arrest & Police arrested someone. & Police used legal authority to catch and take someone to a place where the person may be accused of a crime. (definition of arrest\tablefootnote[13]{\url{https://dictionary.cambridge.org/dictionary/english/arrest}}) \\
\hline
fail & Police failed to intervene. & Police failed to have an effect. (definition of act\tablefootnote[14]{\label{footnote:act}\url{https://dictionary.cambridge.org/dictionary/english/act}}) \\ 
\hline
force & Police used violence. & Police used actions or words that are intended to hurt people. (definition of violence\tablefootnote[15]{\url{https://dictionary.cambridge.org/dictionary/english/violence}}) \\ 
\hline
any action & Police did something. & Police have an effect. (definition of act) \\
\hline
\end{tabular}
\caption{Queries for the India Police Events dataset.}
\label{tab:indiaQuery}
\end{table*}

\subsection{Queries}
We have experimented with different types of queries: definitional queries, extended definitional queries and declarative queries. 

We used the Cambridge Dictionary\footnote{\url{https://dictionary.cambridge.org}} and form the definitional queries by using the definitions of the class name (protest, kill, arrest, etc.). Annotation manual may possibly be a good resource to find the definition of the investigated class. For this reason, we also experimented with definitions from Annotation Manual\footnote{\url{https://github.com/emerging-welfare/general_info/tree/master/annotation-manuals}} \citep{https://doi.org/10.48550/arxiv.2206.10299}. On the other hand, a declarative query is a sentence that simply describes the class. For instance, we use “There is a protest.” as the declarative query for the ProtestNews dataset. For the India Police Events dataset, we use declarative queries proposed by \citet{halterman-etal-2021-corpus}.

We also extended protest dictionary definition by concatenating it with the definitions of words that pass in the query (see last 3 rows in Table \ref{tab:protestQuery}). In one of the queries, the ‘protest’ word is added to the beginning of the protest definition. In the other one, definition of opposition is concatenated with the protest definition. In the third one, definitions of protest and definition of disapproval are concatenated and used as a query. Note that we used the definitions of opposition and disapproval since they occur in the protest definition. All queries used for both datasets are listed in Table \ref{tab:protestQuery} and \ref{tab:indiaQuery}.

\section{Experiments \& Results} \label{experiments}

\begin{figure*}[h]
\begin{subfigure}{.5\textwidth}
  \centering
  \includegraphics[width=.8\linewidth]{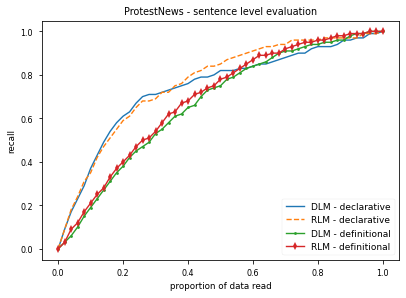}
  \caption{Sentence level evaluation.}
  \label{fig:protest-sent}
\end{subfigure}%
\begin{subfigure}{.5\textwidth}
  \centering
  \includegraphics[width=.8\linewidth]{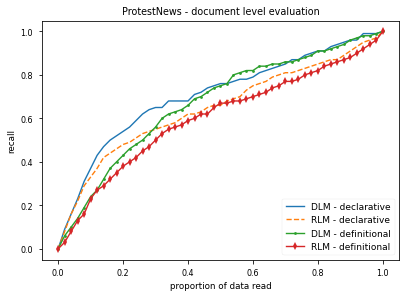}
  \caption{Document level evaluation.}
  \label{fig:protest-doc}
\end{subfigure}
\begin{subfigure}{.5\textwidth}
  \centering
  \includegraphics[width=.8\linewidth]{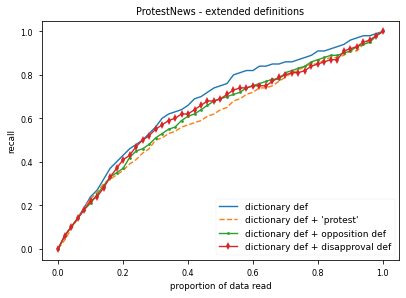}
  \caption{Extended definitions.}
  \label{fig:protest-extended}
\end{subfigure}
\begin{subfigure}{.5\textwidth}
  \centering
  \includegraphics[width=.8\linewidth]{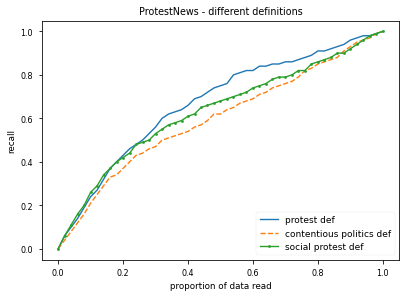}
  \caption{Different definitions.}
  \label{fig:protest-different}
\end{subfigure}
\label{fig:protest}
\caption{ProtestNews dataset tested on two models: RLM and DLM.}
\end{figure*}

\paragraph{ProtestNews Dataset}
is tested with declarative queries, definitional queries and extended definitions on models DLM and RLM and results are presented in Figure \ref{fig:protest-sent} for the sentence level evaluation. The x-axis represents what percentage of the data is read and the y-axis represents how much recall is achieved at that stage. One can investigate what percentage of the data should be read to achieve a specified recall. We see that both models yield similar results when the same query is given but positive documents are accumulated at more top with the declarative query compared to the definitional query.

For document level evaluation, Figure \ref{fig:protest-doc} illustrates the comparison of the models. DLM achieves higher recall scores than RLM, however, the query type does not affect the performance of the model at the document level significantly.

We compare the extended and Annotation Manual definitions at document level using the DLM model since the DLM achieves higher recall compared to RLM at the document level as in Figure \ref{fig:protest-doc}. However, from Figure \ref{fig:protest-extended} we see that extending the protest definition performs slightly worse than using the only dictionary definition. Also, Annotation Manual definitions do not perform better than the dictionary definition as we see from Figure \ref{fig:protest-different}.

\begin{table*}[h!]
\centering
\begin{tabular}{ |p{1.25cm}|p{1.5cm}|c|p{1.25cm} p{1.25cm} p{1.25cm} p{1.25cm} c|  }
 \hline
 \multicolumn{2}{|c|}{} &  \textbf{ProtestNews}& 
 \multicolumn{5}{|c|}{\textbf{India Police Events}}\\
 \cline{3-8}

\multicolumn{2}{|c|}{} & - & kill & arrest & fail & force & any action\\
\hline
DLM & decl-sent & 0.64 &\textbf{0.96 }&\textbf{0.94} &\textbf{0.65} & \textbf{0.91} & \textbf{0.89}\\
\hline
DLM & decl-doc & 0.60 & 0.80 & 0.75 & 0.25 & 0.75 & 0.80\\
\hline
DLM & def-sent & 0.35 & 0.89 & 0.63 & 0.47 & 0.71 & 0.69\\
\hline
DLM & def-doc & 0.41 & 0.62 & 0.42 & 0.21 & 0.21 & 0.65\\
\hline
RLM & decl-sent & \textbf{0.65 }& 0.55 & 0.91 & 0.34 & 0.66 & 0.42\\
\hline
RLM & decl-doc & 0.51 & 0.18 & 0.44 & 0.18 & 0.27 & 0.36\\
\hline
RLM & def-sent & 0.38 & 0.36 & 0.26 & 0.23 & 0.18 & 0.38\\
\hline
RLM & def-doc & 0.34 & 0.11 & 0.15 & 0.16 & 0.11 & 0.37\\
\hline
 
 \end{tabular}
\caption{mAP scores for DLM and RLM models with different document lengths and queries.}
\label{tab:mapScores}
\end{table*} 

\begin{figure*}[h!]
\begin{subfigure}{.5\textwidth}
  \centering
  \includegraphics[width=.8\linewidth]{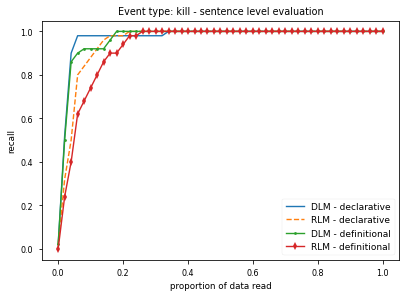}
  \caption{kill}
  \label{fig:kill-sent}
\end{subfigure}%
\begin{subfigure}{.5\textwidth}
  \centering
  \includegraphics[width=.8\linewidth]{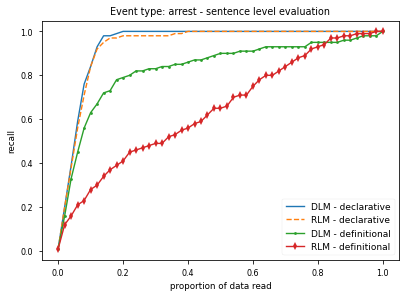}
  \caption{arrest}
  \label{fig:arrest-sent}
\end{subfigure}
\begin{subfigure}{.5\textwidth}
  \centering
  \includegraphics[width=.8\linewidth]{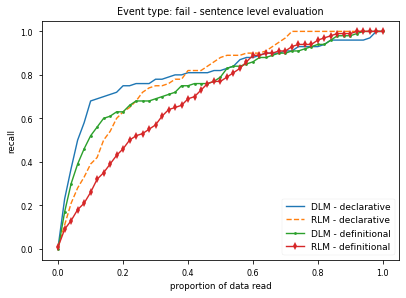}
  \caption{fail}
  \label{fig:fail-sent}
\end{subfigure}
\begin{subfigure}{.5\textwidth}
  \centering
  \includegraphics[width=.8\linewidth]{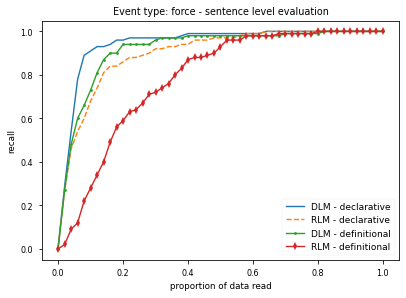}
  \caption{force}
  \label{fig:force-sent}
\end{subfigure}
\begin{subfigure}{.5\textwidth}
  \centering
  \includegraphics[width=.8\linewidth]{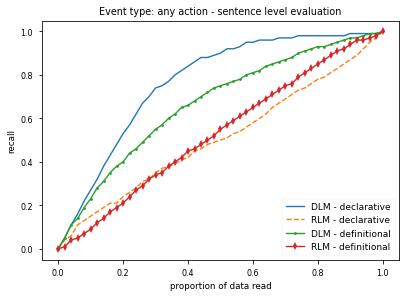}
  \caption{any action}
  \label{fig:anyaction-sent}
\end{subfigure}
\caption{India Police Events dataset sentence level evaluation tested on RLM and DLM.}
\label{fig:india-sent}
\end{figure*}

\begin{figure*}[h!]
\begin{subfigure}{.5\textwidth}
  \centering
  \includegraphics[width=.8\linewidth]{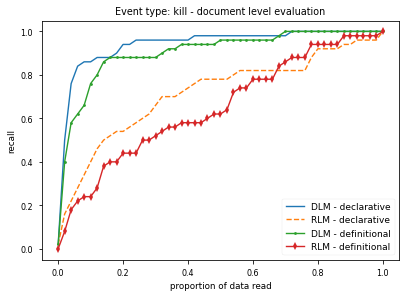}
  \caption{kill}
  \label{fig:kill-doc}
\end{subfigure}%
\begin{subfigure}{.5\textwidth}
  \centering
  \includegraphics[width=.8\linewidth]{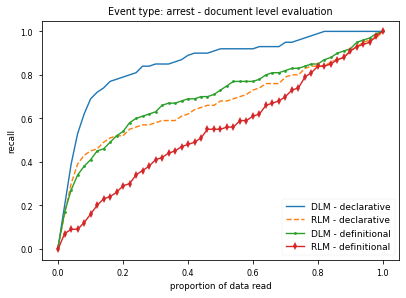}
  \caption{arrest}
  \label{fig:arrest-doc}
\end{subfigure}
\begin{subfigure}{.5\textwidth}
  \centering
  \includegraphics[width=.8\linewidth]{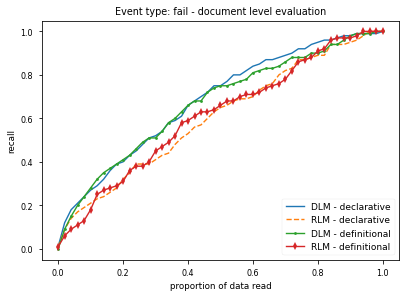}
  \caption{fail}
  \label{fig:fail-doc}
\end{subfigure}
\begin{subfigure}{.5\textwidth}
  \centering
  \includegraphics[width=.8\linewidth]{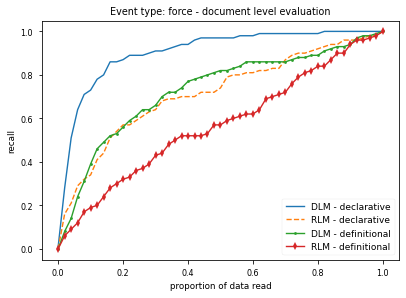}
  \caption{force}
  \label{fig:force-doc}
\end{subfigure}
\begin{subfigure}{.5\textwidth}
  \centering
  \includegraphics[width=.8\linewidth]{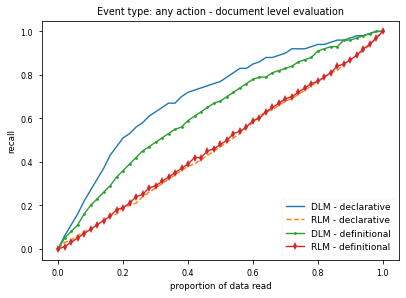}
  \caption{any action}
  \label{fig:anyaction-doc}
\end{subfigure}
\caption{India Police Events dataset document level evaluation tested on RLM and DLM.}
\label{fig:india-doc}
\end{figure*}

\paragraph{India Police Events Dataset}
is tested with declarative and definitional queries on RLM and DLM as in ProtestNews dataset. For all event types, we see from Figure \ref{fig:india-sent} and Figure \ref{fig:india-doc} that DLM with declarative query gives the best result that is positive documents are accumulated at more top-level, whereas RLM with a definitional query stays behind other combinations of model and queries.


\paragraph{Mean Average Precision (mAP)}
is calculated for each ranking and reported in Table \ref{tab:mapScores}. Query and document length combination that gives the highest mAP is marked in bold for each dataset and event type.

For the ProtestNews dataset we observe that using models DLM or RLM, and document lengths do not differ significantly. Whereas using a declarative query gives much better mAP than the definitional query. For the India Police Events dataset for all event types DLM and declarative query with the sentence level evaluation yield the highest score rather than the definitional or document level evaluation. Besides, note that there is a large difference with the other combinations. For example for event type force, sentence level evaluation with DLM and the declarative query gives 0.91 mAP whereas document level evaluation with RLM and the definitional query yields 0.11 mAP.\\
As the topic gets broader, we see that performance gets worse in Table \ref{tab:mapScores}. For instance, kill is a more specific topic than any action since any action event type also includes kill events. When 20\% of the data read, 90\% recall is achieved for event type kill, on the other hand, even 60\% recall is not reached for any action.\\
We take the average sentence and document level mAP scores for each model and present in Table \ref{tab:avgMapScores}. For ProtestNews dataset, sentence or document level does not differ in mAP when DLM is used. However, for India Police Events dataset sentence level evaluation achieves much higher mAP than document level evaluation (0.24 mAP increase for DLM and 0.21 increase for RLM). For both sentence and document level, DLM reaches higher mAP than RLM.
\begin{table}[h!]
\centering
\begin{tabular}{|c|c c|c c|}
 \hline
 \multirow{2}{*}{} &  \multicolumn{2}{c|}{\textbf{ProtestNews}} &  \multicolumn{2}{c|}{\textbf{India Police Events}}\\ \cline{2-5}
& DLM & RLM & DLM & RLM\\
\hline
sent & \textbf{0.50} & \textbf{0.52} & \textbf{0.77} & \textbf{0.43}\\
doc & \textbf{0.50} & 0.42 & 0.53 & 0.22\\
\hline
\end{tabular}
\caption{Average mAP on ProtestNews and India Police Events Dataset for all event types.}
\label{tab:avgMapScores}
\end{table} 
\section{Conclusion} \label{conclusion}
We investigate the performances of two Transformer Language Models (DLM and RLM), different query types (declarative and definitional) in different document lengths (document and sentence level). Our experiments that conclude DLM achieves higher mAP scores than RLM are consistent with the findings of \citet{9673585} and \citet{https://doi.org/10.48550/arxiv.2006.03654}. In general, we find that the combination of DLM with a declarative query in sentence level outperforms other combinations in mAP score. However, scores decrease as the topic or event type gets broader where protest events can be considered broader than specific police actions.

\section{Future Work} \label{future-work}
We plan to analyze results more for example by considering subcategories of protest events for the ProtestNews dataset. Future work can extend this work to a different political event classification dataset and further investigate the association between the broadness of the topic and metric scores. Experiments in languages other than English are also left as future work.

\bibliography{anthology,custom}
\bibliographystyle{acl_natbib}

\appendix

\end{document}